%% file: main.tex
\documentclass[sigconf,natbib=false]{acmart}

\usepackage{tikz}
\usetikzlibrary{external}
\AtBeginDocument{%
  }

\setcopyright{acmlicensed}
\copyrightyear{2025}
\acmYear{2025}
\acmDOI{XXXXXXX.XXXXXXX}
\acmConference[Conference’17]{XXXX}{July 2017}{Washington, DC, USA}
\acmISBN{978-1-4503-XXXX-X/2018/06}

\RequirePackage[
  datamodel=acmdatamodel,
  style=acmnumeric,
  ]{biblatex}
\usepackage{enumitem}
\usepackage{amsmath}
\usepackage{bm}
\usepackage{algorithm}
\usepackage{algpseudocode}

\begin{document}

\title{An Efficient Embedding Based Ad Retrieval with GPU-Powered Feature Interaction}

\settopmatter{authorsperrow=3}
\author{Yifan Lei}
\authornote{Both authors contributed equally to this research.}
\orcid{1234-5678-9012}
\author{Jiahua Luo}
\authornotemark[1]
\authornote{Corresponding author.}
\email{{yifanlei, kennyjhluo}@tencent.com}
\affiliation{%
  \institution{Tencent Inc.}
  \country{Shenzhen, Guangdong, China}
}
\author{Tingyu Jiang}
\email{travisjiang@tencent.com}
\affiliation{%
  \institution{Tencent Inc.}
  \country{Shenzhen, Guangdong, China}
}

\author{Bo Zhang}
\email{peanutzhang@tencent.com}
\affiliation{%
  \institution{Tencent Inc.}
  \country{Shenzhen, Guangdong, China}
}

\author{Lifeng Wang}
\email{fandywang@tencent.com}
\affiliation{%
  \institution{Tencent Inc.}
  \country{Shenzhen, Guangdong, China}
}

\author{Dapeng Liu}
\email{rocliu@tencent.com}
\affiliation{%
  \institution{Tencent Inc.}
  \country{Shenzhen, Guangdong, China}
}

\author{Zhaoren Wu}
\email{jasonzrwu@tencent.com}
\affiliation{%
  \institution{Tencent Inc.}
  \country{Shenzhen, Guangdong, China}
}

\author{Haijie Gu}
\email{jerrickgu@tencent.com}
\affiliation{%
  \institution{Tencent Inc.}
  \country{Shenzhen, Guangdong, China}
}

\author{Huan Yu}
\email{huanyu@tencent.com}
\affiliation{%
  \institution{Tencent Inc.}
  \country{Shenzhen, Guangdong, China}
}

\author{Jie Jiang}
\email{zeus@tencent.com}
\affiliation{%
  \institution{Tencent Inc.}
  \country{Shenzhen, Guangdong, China}
}
\renewcommand{\shortauthors}{Yifan Lei, Jiahua Luo et al.}

\input{abstract}




\maketitle

\input{introduction}
\input{related_work}
\input{method}
\input{experiment}
\input{conclusion}
\printbibliography
\end{document}

%% file: abstract.tex
\begin{abstract}

In large-scale advertising recommendation systems, retrieval serves as a critical component, aiming to efficiently select a subset of candidate ads relevant to user behaviors from a massive ad inventory for subsequent ranking and recommendation. The Embedding-Based Retrieval (EBR) methods modeled by the dual-tower network are widely used in the industry to maintain both retrieval efficiency and accuracy. However, the dual-tower model has significant limitations: the embeddings of users and ads interact only at the final inner product computation, resulting in insufficient feature interaction capabilities. Although DNN-based models with both user and ad as input features, allowing for early-stage interaction between these features, are introduced in the ranking stage to mitigate this issue, they are computationally infeasible for the retrieval stage.

To bridge this gap, this paper proposes an efficient GPU-based feature interaction for the dual-tower network to significantly improve retrieval accuracy while substantially reducing computational costs. Specifically, we introduce a novel compressed inverted list designed for GPU acceleration, enabling efficient feature interaction computation at scale. In other words, we implemented the Wide \& Deep model architecture in the retrieval stage. To the best of our knowledge, this is the first framework in the industry to successfully implement Wide \& Deep in a retrieval system. We apply this model to the real-world business scenarios in Tencent Advertising, and experimental results demonstrate that our method outperforms existing approaches in offline evaluation and has been successfully deployed to Tencent's advertising recommendation system, delivering significant online performance gains. This improvement not only validates the effectiveness of the proposed method, but also provides new practical guidance for optimizing large-scale ad retrieval systems.
\end{abstract}

%% file: introduction.tex
\section{Introduction}
Online advertising recommendation systems are essential tools that leverage user data to deliver personalized ad content, enhancing user engagement and maximizing advertiser returns. These systems have become crucial in powering large-scale online advertising platforms, driving business growth while ensuring tailored user experiences.

A typical modern advertising recommendation system usually has a multi-stage funnel architecture: retrieval, pre-ranking, and ranking stages \cite{covington2016deep, zhang2019deep, wang2020cold}. As the first stage, retrieval rapidly scans through millions of candidate ads to generate a manageable subset of high-potential ads.
This stage relies on extremely efficient models to deal with the whole advertising database.
A prevalent approach is the dual-tower architecture, where user and ad features are independently encoded via deep neural networks (DNNs) to generate embeddings, followed by an inner product operation to compute relevance scores. 

\begin{figure}[ht]
    \centering
    \includegraphics[width=0.43\textwidth]{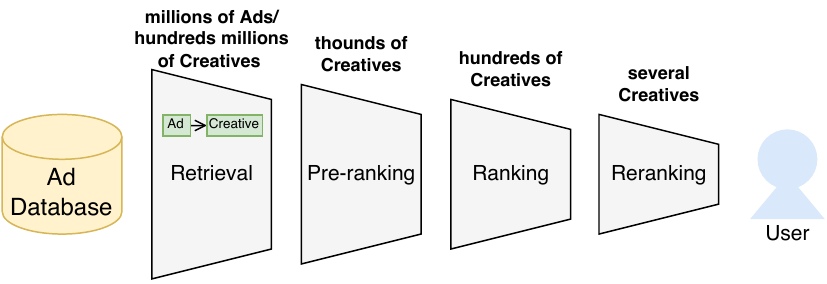}%
    \caption{The overview of the multi-stage funnel architecture for industrial advertising recommendation systems}%
    \label{fig:funnel}%
\end{figure}

For instance, in Tencent's advertising system, retrieval rapidly scans millions of ads to generate a candidate subset of ads using a dual-tower Embedding-Based Retrieval (EBR) \cite{huang2013learning, lin2024enhancing} with an Approximate Nearest Neighbor (ANN) search library.
In the pre-ranking and ranking stages, the selected candidate ads are further expanded into respective creative formats and evaluated using more compute-intensive models to select the optimal candidate ads for user display. This hierarchical funnel framework is illustrated in Figure \ref{fig:funnel}.

The dual-tower model, however, cannot model the interaction between the embeddings of users and ads until the final inner product computation, which leads to insufficient feature interaction capabilities \cite{he2017neural}. Although advanced models such as Inner Product Neural Network (IPNN), Wide \& Deep, DCN and Cold \cite{qu2016product, guo2017deepfm, cheng2016wide, li2024dcnv3, wang2020cold} have been developed for the pre-ranking and ranking stages to capture complex feature interactions, their substantial computational cost makes them infeasible for the efficiency constraints of retrieval tasks.
Recently, many DNN-based retrieval models have been proposed \cite{zhu2018learning, gao2020deep, chen2022approximate}. As they use a DNN scorer for retrieval, these models, however, still suffer from significant query time increases during the retrieval stage, which may further reduce the time available for pre-ranking or ranking, potentially diminishing the overall system's effectiveness.

To address these issues, we propose a retrieval model integrating both implicit and explicit feature interactions through a novel GPU-accelerated operator.
Specifically, we investigate the IPNN \cite{qu2016product, guo2017deepfm} and Wide \& Deep \cite{cheng2016wide} models to the retrieval stage.
While the IPNN emphasizes implicit feature interactions, the Wide \& Deep model specializes in capturing explicit interactions, making their combination well-suited for joint modeling of nuanced user-ad relationships.
Despite their wide usage in later stages of the funnel in pre-ranking and ranking models, these two models have rarely been applied in retrieval due to their computational expense.

To integrate these feature interaction models into the retrieval stage and efficiently handle millions of ads during the retrieval stage, we introduce two strategies for implicit and explicit feature interaction.
For implicit feature interaction, we adopt the dimension extension method to concatenate the implicit feature expression into the original user and item embedding, making it still a dual-tower model.
For explicit feature interaction, we propose a novel compressed inverted list designed for GPU acceleration.
Unlike conventional sparse matrix operations, our method makes use of the binary nature of the feature interaction matrix and organizes ad-feature interactions into GPU-friendly inverted lists.
Although this introduces initial pre-processing overhead, since the index remains constant for a period of time, the cost can be amortized across repeated queries.
With two key considerations, minimizing data transferring between devices and ensuring load balance, our approach organizes data into compact, query-optimized structures, significantly reducing query latency by up to 7x compared to sparse matrix multiplication methods such as the cuSPARSE \cite{naumov2010cusparse} library, a GPU implementation of the sparse matrix multiplication provided by Nvidia.
In summary, the contributions of our paper are as follows:
\begin{itemize}
\item We integrate both implicit and explicit feature interactions into an industry-scale retrieval model. With consideration of the computational overhead of each component in the model, our design balances computational efficiency and model expressiveness, enabling further optimization of complex feature interactions operators.
\item Combining the characteristics of operators, we propose a GPU-powered inverted list index, which significantly accelerates the computation of explicit feature interaction, making its application feasible in retrieval scenarios.
\item Our extensive experiments on real-world commercial datasets show that the proposed model does improve the revenue of the ad recommendation, and the proposed GPU-powered inverted list index can outperform the baseline GPU sparse matrix multiplication method by a large margin.
\end{itemize}


%% file: related_work.tex
\section{Related Work}

\textit{\textbf{Dual-tower Network}}
In advertising recommendation systems,  retrieval stage aims to retrieve top k relevant ads from a large pool of candidates for the subsequent ranking, requiring models that are computationally lightweight, efficient, high-throughput, and low-latency. A prevalent approach in industry is the dual-tower architecture, where user and advertisement features are independently encoded via deep neural networks (DNNs) to generate embeddings, followed by an inner product operation to compute relevance scores. 
Commonly used dual-tower models include DSSM and its variants (e.g., CDSSM\cite{shen2014latent}, MV-DSSM\cite{elkahky2015multi}). 
DSSM employs deep neural networks to map query and document features into a shared semantic space for relevance matching. 
CDSSM extends DSSM by incorporating convolutional neural networks to capture local n-gram patterns in text for enhanced semantic representation. 
MV-DSSM extends the DSSM architecture by incorporating multiple views of data (e.g., text, images, and metadata) into separate towers, enabling joint learning of heterogeneous features for improved semantic matching. 
YouTube DNN \cite{covington2016deep} is designed for recommendation systems, where one tower encodes user features and the other encodes item IDs using a softmax activation function, optimized for large-scale candidate retrieval.
SARec\cite{kang2018self} Integrates self-attention mechanisms into the dual-tower framework to model sequential user behavior, capturing long-range dependencies and temporal patterns for more accurate recommendations.  
BERT4Rec\cite{sun2019bert4rec} is a transformer-based sequential recommendation model that leverages bidirectional self-attention and masked language model pre-training to capture contextual information in user behavior sequences, enhancing recommendation performance.
The dual-tower model retrieve the top ads by maximizing the relevance scores to the user, which is known as Nearest Neighbor Search problem.
To address the Nearest Neighbor Search problem efficiently, Approximate Nearest Neighbor (ANN) techniques are widely adopted, including Locality-Sensitive Hashing (LSH) \cite{har2012approximate, lei2020locality}, Inverted File Index (IVF) \cite{johnson2019billion}, Hierarchical Navigable Small World (HNSW) \cite{malkov2018efficient}, and Product Quantization (PQ) \cite{jegou2010product, avq_2020}, which trade a slight reduction in precision for significant efficiency gains. Clustering-based methods optimize retrieval by narrowing the search space, while graph-based approaches like HNSW use graph traversal for efficient high-dimensional retrieval. Quantization techniques like PQ reduce storage and computational overhead by compressing vector representations.
ANN methods, particularly HNSW and IVF-PQ, are standard for large-scale vector retrieval due to their scalability and efficiency.


\textit{\textbf{Feature Interaction}}
In advertising recommendation systems, feature interaction algorithms are crucial for capturing feature relationships and enhancing model performance, particularly during the fine-ranking stage with small candidate sets, enabling precise ad ranking and scoring. The following are key algorithms in this domain: Factorization Machines (FM) \cite{rendle2010factorization} model second-order feature interactions via latent vector inner products, excelling in sparse data and widely used in recommendations and CTR prediction. Field-aware Factorization Machines (FFM) \cite{juan2016field} extend FM by learning field-specific latent vectors, improving multi-field feature interaction modeling. Field-weighted Factorization Machines (FwFM) \cite{pan2018field} introduces field interaction weights, learning the adaptive interaction weight among feature fields.  Wide \& Deep integrates explicit feature interactions with deep neural networks, designed to achieve a balance between memorization and generalization capabilities. Deep Cross Network (DCN) \cite{wang2017deep} combines deep networks with explicit interaction layers, learning high-order interactions while preserving low-order features. DeepFM integrates FM’s second-order interactions with deep networks for end-to-end training, balancing expressiveness and efficiency. Inner Product-based Neural Network (IPNN) explicitly models interactions via inner products, capturing nonlinear feature combinations. Co-Action Network (CAN) \cite{bian2022can} emphasizes feature pair synergies, flexibly modeling complex relationships in high-dimensional sparse data. 


\textit{\textbf{Inverted List}}
The inverted list data structure is widely used in applications such as large-scale information retrieval and recommendation systems \cite{cutting1989optimization, barroso2003web}.
To reduce storage overhead and enhance query performance, numerous optimization techniques have been developed for inverted list indices. For instance, Variable Byte (VB) and its variants \cite{cutting1989optimization, dean2009challenges} employ a null suppression schema to compress the posting list. VB \cite{cutting1989optimization} leverages 7 bits in a byte to represent the data, with 1 bit indicating whether the current bit is the end of the current integer. Group VB \cite{dean2009challenges} compresses 4 bytes at the same time and uses 2 bits to indicate the number of valid bytes per integer. PforDelta \cite{zukowski2006super} proposes to compress the delta code of each list. The delta code of each integer is partitioned into blocks, and in each block, the smallest possible bit length $b$ is chosen to represent the values in the block in $b$-bits. To improve the query performance of inverted lists, SIMDPforDelta \cite{lemire2015decoding}, as a SIMD version of PforDelta, utilizes SIMD instructions to accelerate the processing of the inverted list index.
In recent years, GPU-accelerated inverted list indices have been proposed to harness the parallel processing capabilities of GPUs \cite{zhou2018generic}. For instance, Zhou et al. \cite{zhou2018generic} demonstrated the use of GPU-accelerated inverted lists for similarity search, enabling efficient parallel processing of batched queries. In \cite{shanbhag2022tile}, Shanbhag et al. proposed to use a cascading decompression schema optimized for GPU architecture to compress the inverted list.

%% file: method.tex
\section{Methodology}

In this section, we introduce the proposed model in detail. First, we introduce the overview of the dual-tower network in the learning to rank framework. Then, we show the proposed implicit and explicit feature interaction modules for the dual-tower model. Finally, we present our proposed GPU inverted list and how it can accelerate the computation of the explicit feature interaction operator.

\subsection{Learning-to-Rank based Dual-Tower Model in Ad Retrieval}
From a mathematical perspective, retrieval can be defined as a preference function that selects a subset of high-scoring items from a candidate set. The scoring function is usually trained using the dual-network, where user and ad features are separately transformed to a low dimensional representation \( \mathbf{h}_u \in \mathbb{R}^d \) and \( \mathbf{h}_a \in \mathbb{R}^d \) respectively.
For a dual-tower model, the preference score \(s= s(u, a) \) of user \( u \) for ad \( a \) is computed as the inner product of the user representation \( \mathbf{h}_u \) and the ad representation \( \mathbf{h}_a \)
\begin{equation}
    s(u, a) = \langle \mathbf{h}_u, \mathbf{h}_a \rangle = \mathbf{h}_u^\top \mathbf{h}_a
    \label{eqn:dual-tower}
\end{equation}
We utilize the LambdaRank framework\cite{burges2010ranknet}  with pairwise logistic loss \cite{liu2009learning}  and non-differentiable measurement metrics to train the model on a dataset where the labels represent the desired ranking outcomes.  Specifically,  the loss function for the ranking function contains two parts, pairwise logistic loss and measurement metrics, as defined as:
\begin{gather}
\mathcal{L}=\sum_{(i,j):a_i<a_j}\log(1+e^{-(s_i-s_j)})\cdot|\Delta\mathrm{NDCG}_{ij}|
\end{gather}
The metric \(\mathrm{NDCG}\) is concerned with the ranking of top candidates, which matches the goal of retrieval—to return a top-N set from millions of candidates. It is formally defined as follows,
\begin{gather}
\mathrm{DCG_i}=\sum_{i=1}^{i=D}\frac{2^{p_i}-1}{log(i+1)} \\
=\sum_{i=1}^{i=D}\frac{2^{D-i}-1}{log(i+1)} \\
\mathrm{NDCG_i}=\frac{DCG_i}{maxDCG} 
\end{gather}
where $D$ is the length of the sampled advertisement sequence.
However, the LambdaRank algorithm, with its core metric $\Delta$NDCG, is originally designed for relevance ranking in document retrieval. It computes ranking cost solely from the relative order of items, ignoring their intrinsic value. This poses a limitation in ad ranking, where eCPM directly determines revenue in ad bidding like oCPM: misordering ads with divergent eCPMs results in greater financial loss than with similar eCPMs. To integrate this business metric, we refine the original LambdaRank loss.

\begin{equation}
\mathcal{L}=\sum_{(i,j):a_i<a_j}\log(1+e^{-(s_i-s_j)})\cdot\left\{|\Delta\mathrm{NDCG}_{ij}|\odot{|\triangle Value_{ij}|}\right\}
\end{equation}
where $\odot$ denotes multiplication or addition operator. $\triangle Value$ can be defined as the loss of the physical value for the ground truth incurred by swapping the positions of ads $a_i$ and $a_j$. Its physical implication is that misranking two ads with a significant commercial value difference leads to a greater penalty.  In practical application, researchers or developers can select the most appropriate non-differentiable loss function in the LambdaRank framework according to specific scenario and data characteristics\cite{burges2010ranknet}.





\begin{figure}[th]
    \centering
    \includegraphics[width=0.45\textwidth]{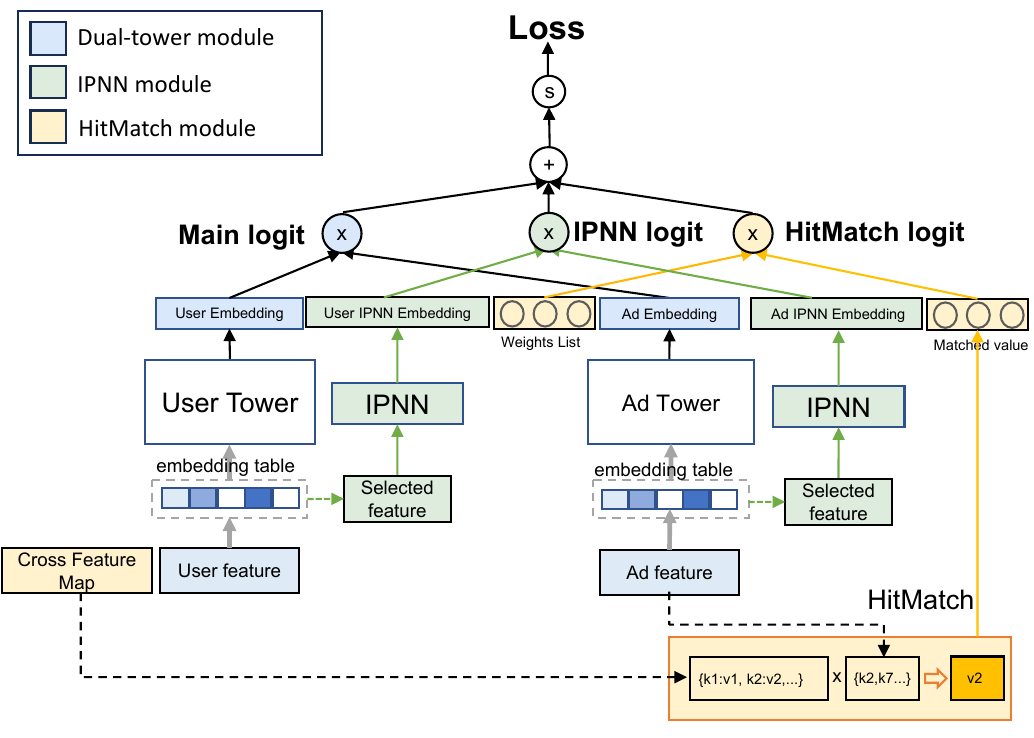}%
\caption{
The proposed model architecture with three core components: a dual-tower network, an IPNN module, and a HitMatch module, which generate the main logit, IPNN logit, and HitMatch logit. These three logits are aggregated through summation at the final layer of the model.}
\label{fig:proposed-model} 
\end{figure}

As the dual-tower model is efficient for retrieval tasks but struggles with complex feature interactions due to its separate architecture of user and ad features. To improve this, two new modules, implicit and explicit feature interaction, are introduced to the dual-tower in the following subsections. The implicit module uses latent space transformations to capture subtle dependencies, while the explicit module models higher-order feature interactions directly. These modules, when integrated into the dual-tower architecture, enhance its expressiveness without sacrificing scalability. The structures of the proposed modules are illustrated in Figure \ref{fig:proposed-model}.

\subsection{Implicit Feature Interaction}
In this subsection, we introduce the IPNN operator designed to capture the interaction between user and ad features.
Let \( \bm{u} = (u_1, u_2, \dots, u_n) \) and \( \bm{v} = (v_1, v_2, \dots, v_m) \) represent the concatenation of a certain user and ad features, respectively, after being transformed through embedding tables and pooling operations. The IPNN operator is decomposed as 
\begin{equation}
\sum\langle u_{i},v_{j}\rangle=\langle \Tilde{\bm{u}},\Tilde{\bm{v}}\rangle;\ 
\Tilde{\bm{u}}=\bm{W^{(u)}} \bm{u};\ 
\Tilde{\bm{v}}=\bm{W^{(v)}} \bm{v},
\end{equation}
where $\bm{W^{(u)}}\in \mathbb{R}^{d\times n}$ and $\bm{W^{(v)}} \in \mathbb{R}^{d \times m}$ denote the weight matrices corresponding to the fields $u$ and $a$, respectively.
Therefore, by applying the IPNN operator to the dual-tower model, we can first compute \( \Tilde{\bm{u}} \) and \( \Tilde{\bm{v}} \) separately on the user and ad, respectively, and then perform the inner product operation at the top layer of the model. In practical applications, by expanding the lengths of the user and ad vectors, we concatenate the summed \( \Tilde{\bm{u}} \) and \( \Tilde{\bm{v}} \) after the \( \bm{h}_u \) and \( \bm{h}_a \) respectively. This model remains as a dual-tower structure, maintaining high computational efficiency.
Specifically, the new logit calculation formula is:
\begin{equation}
s(u, a) = \Tilde{\bm{h}}_u \Tilde{\bm{h}}_a = [\bm{h}_u, \Tilde{\bm{u}}]^T [\bm{h}_a, \Tilde{\bm{v}}],
\end{equation}
where $[\cdot, \cdot]$ denotes the vector concatenation. 
\subsection{Explicit Feature Interaction}
The integration of the IPNN operator can improve the feature interaction capabilities for the dual-tower architecture to some extent. However, its structure remains inherently a dual-tower network which still exhibits limitations in feature interaction. In this subsection, we investigate a lightweight, non-dual-tower architecture, inspired from the widely-adopted Wide \& Deep model in the ranking phase. Here we adapt it to the retrieval phase and name it HitMatch module throughout this paper. The proposed module is depicted in Figure \ref{fig:proposed-model}, highlighted with yellow color.
To minimize computational overhead, the wide part features are selected cautiously, only including their side information and the IDs of ads that users have historically interacted with. The side information includes multi-level ad categories, tags, advertiser IDs, and item IDs, etc. These historical behavior features are grouped as multiple sequences associated with the user ID.
During the training process, the ad component in the side information feature looks up the matched feature values from the user sequences. It then performs a weighted sum using the associated learnable weights to generate the logit for the cross part.

Suppose that there are $N$ ads and the wide component contains \(M\) cross features, where the associated weights are \(\textbf{w} = \{w_1, w_2, \dots, w_M\}\) and their values are \(\textbf{x} = \{x_1, x_2, \dots, x_M\}\).
We denote whether the feature values empty or not by $\bm{L} \in \{0, 1\}^{N \times M}$, where each element $\bm{L}_{a, i}=1$ when ad $a$ whose feature-$i$ not empty, and otherwise $\bm{L}_{a, i}=0$. $\bm{L}$ is highly sparse for a specific ad.
 The final score of ad \( a \) is calculated as
\begin{equation}
s(u, a) = \langle \Tilde{\bm{h}}_u, \Tilde{\bm{h}}_a \rangle +\sum\limits_{i=1}^{M}{w_i x_i \bm{L}_{a, i}}
\label{eqn:hitmatch}
\end{equation}
Here, \(\Tilde{\bm{h}}_u\) and \(\Tilde{\bm{h}}_a\) correspond to the concatenated outputs of the user tower and the ad tower, respectively, derived from both the dual-tower structure and the IPNN module. In the inference phase, the inner product computation in the dual-tower part is very lightweight. However, the HitMatch operator is computationally expensive due to the large number of candidate ads, which can reach millions. Therefore, it is necessary to carefully design an efficient indexing operator to quickly compute results within a short time frame. In the next section, we will introduce the efficient indexing HitMatch operator that we have developed.

\subsection{GPU Inverted List based HitMatch Operator}
\begin{figure*}[ht]
    \centering
    \includegraphics[width=0.96\textwidth]{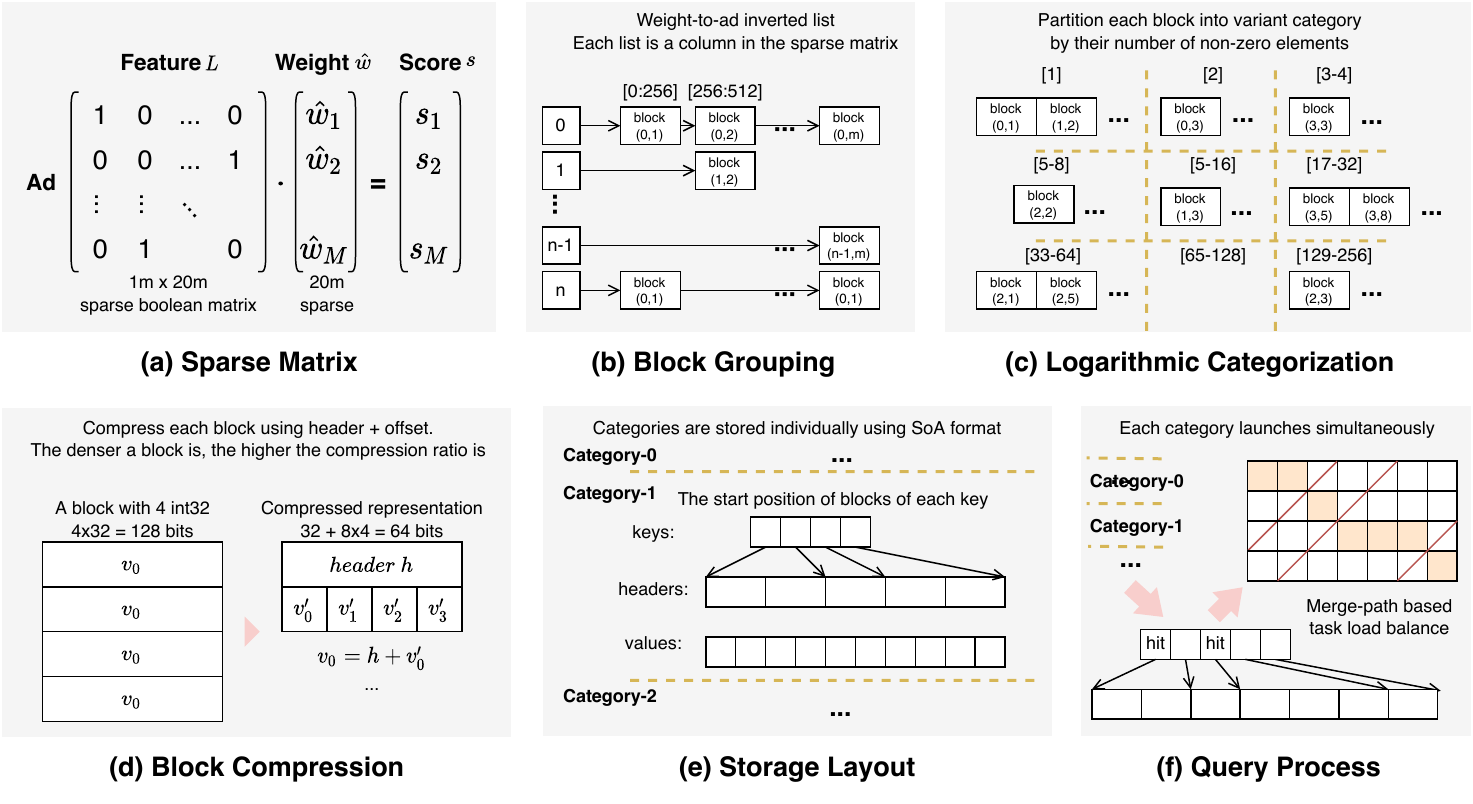}%
    \caption{The overall proceed of the proposed GPU-powered HitMatch operator}%
    \label{fig:hitmatch} 
\end{figure*}

As shown in Equation \ref{eqn:hitmatch}, the HitMatch score can be calculated using a sparse matrix multiplication $\bm{L} \bm{\Tilde{w}}$, where $\Tilde{w}_i=w_i x_i, \forall i$.
Computing this sparse matrix multiplication, however, is still much slower than computing the dual-tower scores with a dense matrix multiplication. This significant performance gap renders a naive application of sparse matrix operations impractical in real-world scenarios requiring high throughput and low latency.

To overcome this limitation, we observe that the sparse matrix $\bm{L}$ consists solely of binary values and remains constant across different queries.
These properties allow us to preprocess $\bm{L}$ into an optimized index prior to any query execution.
Specifically, we recognize the binary matrix $\bm{L}$ is naturally in the form of an inverted list. Furthermore inverted lists are widely used in various fields and are well-suited for GPU processing \cite{zhou2018generic, li2023slim, bruch2024efficient}.
Consequently, we propose a novel GPU-powered inverted list data structure to preprocess the matrix $\bm{L}$.

Our design is driven by two key considerations: (a) \textbf{Load balance:} the computation is done on GPU, so the load balance is essential for the performance, and (b) \textbf{Reducing IO:} the sparse matrix multiplication or inverted list reduction is considered as an IO-bound operator, so reducing the size of data to be processed is another aspect to achieve high performance.
With these two considerations, we represent the matrix $\bm{L}$ as an compressed inverted list, where each list represents a column in $\bm{L}$, the keys are the feature indices, and the values are the ad indices.
The processing proceeds are shown as follows:


\begin{itemize}[leftmargin=0pt,label=]
\item \textbf{Block Grouping}: To ensure efficient storage and computation, we first split lists into blocks by the lowest 8 bits of ad indices and consider each block as a basic unit of the index in the following process.
This ensures uniform block sizes, which make it possible to further compress the data storage. 
\item \textbf{Logarithmic Categorization}: Next, we partition blocks into variant groups according to their number of ads in each block. For a block containing $n$ ads, it will be partitioned into the $\lceil \log_2(n) \rceil$-th group. Then, we pad each block into the same length of $2^{\lceil \log_2(n) \rceil}$. Hence, the block in the same group will have the similar workload, which is essential for the load-balance.
\item \textbf{Block Compression}: Once the blocks are grouped, we compress each block to minimize storage requirements. Specifically, for each block, we utilize only one header value representing the same highest 24 bits for all the ads in the block so we can store an ad in a block by only extra 8 bits integers by their lower 8-bits. This minimizes the size of data storage, significantly alleviating IO bottlenecks.
\item \textbf{Storage Layout}: We store the blocks in a struct-of-arrays (SoA) format, enhancing memory coalescing during GPU computation \cite{baghsorkhi2012efficient}. In order to locate the begin and end of the data given each key efficiently, for keys array, we store the segment of each key indicating the offset of the each key.
\end{itemize}

The overall processing pipeline of overall indexing procedure is visualized in Figure \ref{fig:hitmatch}, and the detailed implementation is outlined in Algorithm \ref{alg:indexing}.

When a query comes, each block group is computed in parallel separately.
For each non-zero feature value $x$ of query, we lookup all the blocks using its index and add the score for each ad by $\Tilde{{w}_i}$.
To further improve the load balance, we integrate the merge-based load balance algorithm \cite{Baxter:2016:M2}. This approach dynamically assigns computational tasks to GPU threads according to the number of blocks within each list for the current block group.
Specifically, our approach first locates the starting and ending positions of target data entries within the stored key array based on the provided keys. We then compute the prefix sum of data lengths per key using a parallel exclusive scan algorithm\cite{merrill2016single} to determine data offsets. Subsequently, a merge-based load balancing strategy dynamically assigns tasks to all GPU threads in parallel, utilizing the precomputed keys and offsets to ensure optimal thread utilization.
By adjusting the workload distribution in real-time, our method further handles imbalances caused by uneven block sizes.
Figure \ref{fig:hitmatch} (d) shows an overview of the HitMatch operator implementation, and the query processing algorithm is formalized in Algorithm \ref{alg:query}.

In our advertising retrieval system, the feature-to-ad mapping matrix updates every few minutes. Upon receiving a matrix update, the server immediately constructs a GPU-accelerated inverted list via Algorithm \ref{alg:indexing}, and use this structure to accelerate the subsequent explicit feature interaction computations. Given that a single built index can serve hundreds of thousands of queries, the associated indexing overhead is effectively amortized due to its reusability across numerous inference requests.

\begin{algorithm}[ht]
\caption{\textsf{Indexing Algorithm}}\label{alg:indexing}
\begin{algorithmic}[1]
\Require Input key-value map $L$ \Comment{keys: feature indices, values: ad indices}
\State $B \gets$ Empty hash map with default value $\emptyset$ \Comment{Blocks}
\State $G \gets$ Empty hash map with default value $\emptyset$ \Comment{Groups}

\ForAll{$(k, v) \in L$ \textbf{in parallel}} \Comment{Step 1: Block Grouping and Compression}
    \State $h, l \gets \lfloor v / 2^8 \rfloor, v \bmod 2^8$
    \State $B[k, h] \gets B[k, h] \cup \{l\}$
\EndFor

\ForAll{$(k, h), ls \in B$ \textbf{in parallel}} \Comment{Step 2: Group Partitioning}
    \State $n \gets |ls|$
    \State $g \gets \lceil \log_2(n) \rceil$ 
    \State Pad $ls$ with $0$ to length $2^g$
    \State $G[g, k, h] \gets G[g, k, h] \cup \{ls\}$
\EndFor

\State Allocate arrays $\textsf{keys}$, $\textsf{header}$, $\textsf{values}$ for each group in $G$ \Comment{Step 3: Struct-of-Array (SoA) Storage}
\For{$g \in \{0, 1, \cdots, 8\}$ \textbf{in parallel}}
    \State $i \gets 0$
    \State Append $0$ to $\textsf{keys[g]}$
    \ForAll{$(k, h), ls \in G[g]$}
        \State Append $h$ to $\textsf{header[g]}$
        \State Append $ls$ to $\textsf{values[g]}$
        \While{ $i < k$}
            \State Append $k$ to $\textsf{keys[g]}$
            \State $i \gets i + 1$
        \EndWhile
    \EndFor
\EndFor

\State \Return $\textsf{keys}, \textsf{header}, \textsf{values}$
\end{algorithmic}
\end{algorithm}

\begin{algorithm}[t]
\caption{\textsf{Query Algorithm}}\label{alg:query}
\begin{algorithmic}[1]
\Require Query $\mathbf{q}$ with non-zero feature values $\{(k, \hat{w})\}$, Indexed data in SoA format: $\{\textsf{keys}, \textsf{header}, \textsf{values}\}$
\State Initialize $scores$ as a zero array of size equal to the number of ads
\For{$g \in \{0, 1, \cdots, 8\}$ \textbf{in parallel}}
    \State $k\_length \gets [\textsf{keys}[g, k_i + 1] - \textsf{keys}[g, k_i]\ for\ k_i \in \mathbf{q}]$
    \State $k\_seg \gets ExclusiveScan(k\_length)$ 
    \ForAll{$k, k\_offset \in LoadBalance(k\_seg)$ in parallel based on \cite{Baxter:2016:M2}}
        \State $h \gets \mathbf{header[g, k\_offset]}$
        \State $ls \gets \mathbf{values[g, k\_offset]}$
        \ForAll{$l \in ls$}
            \State $AtomicAdd(\textsf{scores}[h \cdot 2^8 + l], \hat{w}[k])$
        \EndFor
    \EndFor
\EndFor
\State \Return $\textsf{scores}$
\end{algorithmic}
\end{algorithm}






%% file: experiment.tex
\section{Experiments}
In this section, we conduct experiments to evaluate our proposed methods, aiming to answer the following key research questions: 
\begin{itemize}[leftmargin=0pt,label=]
\item\textbf{RQ1:} Does our method achieve superior offline state-of-the-art (SOTA) performance compared to standard baseline models?  
\item\textbf{RQ2:} Can our method significantly enhance online advertising revenue?  
\item\textbf{RQ3:} Does our method demonstrate more efficient resource utilization in terms of engineering performance compared to existing sparse matrix multiplication methods?  
\end{itemize}

\subsection{Experimental Setup}
\subsubsection{Datasets}
We collect historical advertising data to evaluate our proposed models from the WeChat-channel video feed, a prominent short video sharing platform within WeChat. The reason why we do not choose publicly available benchmark in this study is due to the critical need for evaluating the proposed GPU Hitmatch operator on  a large-scale advertisement candidate pool. To meet these demands,  we gathered seven days of  ad request logs to evaluate our proposed algorithms.  The first six days' logs were used for training dataset, while the seventh day’s data were reserved for testing dataset. To measure the engineering performance metrics of the model’s inference capabilities, we utilized the online advertisement candidate pool containing millions of instances, enabling a thorough evaluation of the system’s scalability and operational efficiency.

\subsubsection{ Features}
Our model leverages user-side features categorized into basic user profiles and historical behavioral data organized in sequence. The former encompasses demographic and the latter records user interactions with the ad videos, including views, follows, likes, clicks, and other engagement metrics. Additionally, the ad features include material attributes (e.g., title, description, images, and videos), hierarchical categories, bidding price, optimization goals and other meta data.


In this study, the cross-features including the hierarchical categories, advertiser and other meta data from the advertisements that users have previously interacted with as the input for the HitMatch operator.  Such features are statistically aggregated dynamically in real time, including the click counts, click-through rates,  conversion counts, and conversion rates. The aggregation time windows are performed in multiple period,such as 1, 7, 15, 30 days and so on, encompassing advertisements that users have engaged with during this period.

\subsubsection{Compared Models and Evaluation Metrics}
In our evaluation, we perform a thorough analysis of various models, including the baseline dual-tower model (DT) and the proposed DT-IPNN-HitMatch. To ensure a comprehensive comparison, we also conduct additional experiments with the dual-tower model integrated with the IPNN module (DT-IPNN) and the HitMatch module (DT-HitMatch).
We mainly evaluate the following metrics: 
\begin{itemize}[leftmargin=0pt,label=]
\item \textbf{GAUC}: This metric computes the weighted average of AUC values at the user level, where the top-1 ad is designated as the positive instance and the remaining as negative.
\begin{equation}
    GAUC=\frac{1}{\#pv}\sum_{i=1}^{\#pv}AUC_i
\end{equation}
where \#pv denotes number of requests.
\item \textbf{Recall@G\_k}: This metric quantifies the proportion of top-k ranked ads in the final ranking that are present within the top-G retrieval ads, with higher values indicating superior performance. As only one ad is displayed per request in our scenario and we only sample 30 ranked ads per request for training model. Thus, 
 we primarily focuses on the case where k = 1,   $G \in [5,10]$ in offline evaluations and $G \in [100]$ in online evaluations.

\begin{equation}
    \mathrm{Recall@}G\_k=\frac{1}{\#pv}\sum_{i=1}^{\#pv}\frac{|\{ TopG\ ads\}\cap\{TopK\ ads\}|}{k}
\end{equation}
\item \textbf{Online metrics}: For online metrics, we mainly care about advertising cost and GMV (Gross Merchandise Volume)
\item \textbf{Efficiency metrics}: Maximum QPS and pre-processing time for our inverted list index for the HitMatch operator.
\end{itemize}

We train the aforementioned models using the LTR paradigm, with sampling 30 ads in the final ranking phase per request. The model is trained online in real-time with a batch size of 4000. User and ad feature embeddings are 32-dimensional. Dense DNN parameters are optimized using Adam (learning rate: 0.015, beta: 1.0), while sparse parameters use FTRL (learning rate: 1.0E-4, beta\_1: 0.5, beta\_2: 0.999, epsilon: 1.0E-8). Both user and ad towers produce 64-dimensional vectors using DNN.
\subsection{Offline Evaluations (RQ1)}

\begin{table}[t]
  \caption{Overall performance comparison on the different methods in offline evaluation}
  \label{tab:offline-result}
  \begin{tabular}{cccc} 
    \toprule
    Method & GAUC & Recall@5\_1& Recall@10\_1\\
    \midrule
    DT      & 0.839& 0.730& 0.908\\
    DT-IPNN    & 0.843          & 0.739& 0.913\\
    DT-HitMatch    & 0.847& 0.753& 0.924\\
    \textbf{DT-IPNN-HitMatch} & \textbf{0.861 }& \textbf{0.768}&\textbf{0.939}\\
    \midrule
    Improvement over DT& \textbf{2.62}\%& \textbf{5.21}\%& \textbf{3.41}\%\\
    \bottomrule
  \end{tabular}
\vspace{-0.5em} 
\end{table}

The offline experimental evaluation results, as summarized in Table \ref{tab:offline-result}, provide a detailed comparison of the performance of our proposed models against the baseline DT model across three key metrics: GAUC, Recall@5\_1, and Recall@10\_1. The results clearly indicate that all three proposed models—DT-IPNN, DT-HitMatch, and DT-IPNN-HitMatch—consistently outperform the baseline DT model, with DT-IPNN-HitMatch achieving the highest overall performance. Specifically, DT-IPNN-HitMatch demonstrates significant improvements, achieving a GAUC of 0.861, a Recall@5\_1 of 0.768, and a Recall@10\_1 of 0.939, which represent relative gains of 2.62\%, 5.21\%, and 3.41\%, respectively, over the baseline DT model.
The superior performance of DT-HitMatch and DT-IPNN-HitMatch compared to DT-IPNN and DT can be attributed to the incorporation of explicit cross-features, which serve as strong signals in our system. These cross-features effectively capture the interactions between user preferences and item characteristics, leading to more accurate recommendations. Additionally, the stability of user interests over a given time period further enhances the effectiveness of these features, as they allow the model to better align recommendations with long-term user preferences.



\subsection{HitMatch Operator Evaluations (RQ2)}
To illustrate the performance of the proposed inverted list index for the explicit feature interaction, we also conduct the experiment against the cuSPARSE \cite{naumov2010cusparse}
library, a GPU implementation of the sparse matrix multiplication provided by Nvidia. We collect the sparse matrix $\bm{L}$ from the historical advertising data and utilize 1000 feature vectors as queries to benchmark the time usage between cuSPARSE and the proposed approach on an instance with a Nvidia Tesla T4 graphic card.
The results are shown in Table \ref{tab:hitmatch}. As can be seen, our method outperforms the cuSPARSE by around 7 times in QPS(+590\%). This shows that the proposed approach indeed have superior performance owing to the lower data transformation and better load balance by introducing a small preprocessing time.
The preprocessing time of our method is 224 ms, which is negligible amortized by a large number of queries.

\begin{table}
  \caption{The benchmark of the explicit feature interaction operation}
  \label{tab:hitmatch}
  \begin{tabular}{ccc}%
    \toprule
    Method & Preprocessing time (ms) & QPS (1/s) \\
    \midrule
    cuSPARSE     & N.A.    & 275.94 \\
    Ours    & 224.32    & \textbf{1904.23} \\
    \midrule
    Improvement & N.A.   & \textbf{590}\%  \\
    \bottomrule
  \end{tabular}
\vspace{-0.5em} 
\end{table}
\subsection{Online Evaluations (RQ3)}
We deployed our proposed models using the aforementioned feature settings in the Tencent Advertising online system with A/B testing. The proposed model demonstrated significant performance improvements as shown in table \ref{tab:online-results}, achieving cost, GMV and Recall@100\_1 increase by 0.37\%, 1.58\% and 1.8\%  respectively in the WeChat Moment.  With 1.25\%, 1.49\% and 2.5\% increase in cost, GMV and Recall@100\_1 respectively in WeChat Channel. It is worth to noting that the performance is closely tied to the design of cross-feature types and their quantity in practical application. Generally, richer and more diverse cross-features may lead to improved results.  For millions of ads in WeChat Moments and Channel, the proposed HitMatch operator can be done within 500 us on a Nvidia T4 GPU, demonstrating its ability to efficiently address explicit feature interaction problem at an industry scale. 

\begin{table}
  \caption{Improvement in online evaluation by A/B Test }
  \label{tab:online-results}
  \begin{tabular}{cccl}%
    \toprule
    Traffic& cost& GMV& Recall@100\_1 \\
    \midrule
    WeChat-Moment& +0.37\%& +1.58\%& +1.8\%\\
    WeChat-Channel& +1.25\%& +1.49\%& +2.5\%\\ \bottomrule
  \end{tabular}
\vspace{-0.5em} 
\end{table}

%% file: conclusion.tex
\section{Conclusion}
In this paper, we propose a novel approach from a model-engineering co-design perspective to integrating both implicit and explicit feature interactions for ad retrieval models, leveraging a GPU-accelerated inverted list index. Our method significantly reduces computational overhead, enabling real-time processing of millions of ads in our platforms such as WeChat Moments and Channel.
Our extensive experiments on real-world datasets demonstrate substantial improvements in recommendation accuracy and query latency, outperforming the baseline by a large margin.
These results demonstrate the effectiveness of our method, and providing a practical insight for optimizing large-scale ad retrieval systems.